\documentclass[10pt,twocolumn,letterpaper]{article}
\usepackage{url}
\usepackage{cvpr}
\usepackage{times}
\usepackage{epsfig}
\usepackage{graphicx}
\usepackage{amsmath}
\usepackage{amssymb}
\usepackage{multirow}
\usepackage{subcaption}
\usepackage{enumitem}
\usepackage{wrapfig}
\usepackage{capt-of,etoolbox}
\usepackage{caption}
\usepackage{lipsum}
\usepackage[ruled,vlined]{algorithm2e}
\usepackage{epstopdf} 

\newcommand\bigfrown[2][\textstyle]{\ensuremath{%
  \array[b]{c}\text{\scalebox{2}{$#1\frown$}}\\[-1.3ex]#1#2\endarray}}
  
\cvprfinalcopy 

\def\cvprPaperID{***} 

\begin{document}

\title{Im2Fit: Fast 3D Model Fitting and Anthropometrics using Single Consumer Depth Camera and Synthetic Data}

\author{Qiaosong Wang$^\dag$, Vignesh Jagadeesh$^\ddag$, Bryan Ressler$^\ddag$, Robinson Piramuthu$^\ddag$\\
$^\dag$University of Delaware $^\ddag$eBay Research Labs\\
{\tt\small qiaosong@udel.edu,[vjagadeesh, bressler, rpiramuthu]@ebay.com}}

\ifcvprfinal\pagestyle{empty}\fi

\twocolumn[{%
\renewcommand\twocolumn[1][]{#1}%
\maketitle
\begin{center}
\centering
\includegraphics[width=\linewidth]{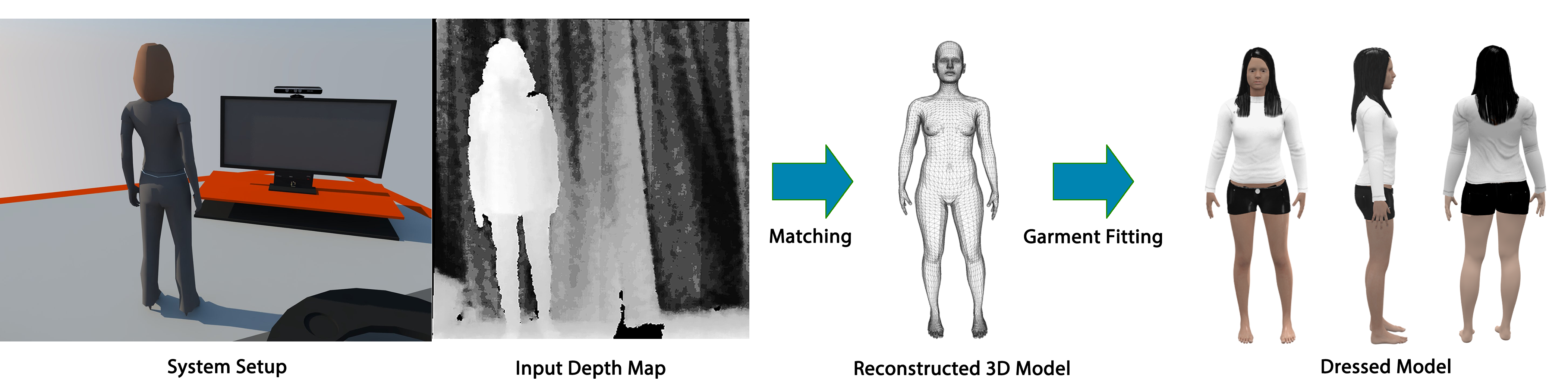}
\captionof{figure}{\textbf{Im2Fit}. We propose a system that uses a single consumer depth camera to (i) estimate 3D shape of human, (ii) estimate key measurements for clothing and (iii) suggest clothing size. Unlike competing approaches that require a complex rig, our system is very simple and fast. It can be used at the convenience of existing set up of living room.}
\label{fig:Teaser}
\end{center}%
}]

\begin{abstract}
   Recent advances in consumer depth sensors have created many opportunities for human body measurement and modeling. Estimation of 3D body shape is particularly useful for fashion e-commerce applications such as virtual try-on or fit personalization. In this paper, we propose a method for capturing accurate human body shape and anthropometrics from a single consumer grade depth sensor. We first generate a large dataset of synthetic 3D human body models using real-world body size distributions. Next, we estimate key body measurements from a single monocular depth image. We combine body measurement estimates with local geometry features around key joint positions to form a robust multi-dimensional feature vector. This allows us to conduct a fast nearest-neighbor search to every sample in the dataset and return the closest one. Compared to existing methods, our approach is able to predict accurate full body parameters from a partial view using measurement parameters learned from the synthetic dataset. Furthermore, our system is capable of generating 3D human mesh models in real-time, which is significantly faster than methods which attempt to model shape and pose deformations. To validate the efficiency and applicability of our system, we collected a dataset that contains frontal and back scans of 83 clothed people with ground truth height and weight. Experiments on  real-world dataset show that the proposed method can achieve real-time performance with competing results achieving an average error of 1.9 cm in estimated measurements.
\end{abstract}



\section{Introduction}
\label{sec:Introduction}
Recent advances in 3D modeling and depth estimation have created many opportunities for non-invasive human body measurement and modeling. Particularly, getting precise size and fit recommendation data from consumer depth cameras such as Microsoft\textregistered~Kinect\texttrademark~device can reduce returns and improve user experience for online fashion shopping. Also, applications like virtual try-on or 3D personal avatars can help shoppers visualize how clothes would look like on them. According to \cite{Apparel12,Fitsme12}, about $\$44.7$B worth of clothing was purchased online in $2013$. Yet, there was an average return rate of $25\%$ because people could not physically try the clothes on. Therefore, an accurate 3D model of the human body is needed to provide accurate sizing and measurement information to guide online fashion shopping. One standard approach incorporates light detection and ranging (LiDAR) data for body scanning \cite{THD,GGS}. An alternative approach is to use a calibrated multi-camera rig and reconstruct the 3D human body model using structure from motion (SfM) and Multi-View Stereo (MVS) \cite{deAguiar}. A more recent work \cite{tsoliWACV14} aims at estimating high quality 3D models by using high resolution 3D human scans during registration process to get statistical model of shape deformations. Such scans in the training set are very limited in variations, expensive and cumbersome to collect and nevertheless do not capture the diversity of the body shape over the entire population. All of these systems are bulky, expensive, and require expertise to operate.

 Recently, consumer grade depth camera has proven practical and quickly progressed into markets. These sensors cost around $\$200$ and can be used conveniently in a living room. Depth sensors can also be integrated into mobile devices such as tablets \cite{SIO}, cellphones \cite{TANGO} and wearable devices \cite{AHL}. Thus, depth data can be obtained from average users and accurate measurements can be estimated. However, it is challenging to produce high-quality 3D human body models, since such sensors only provide a low-resolution depth map (typically 320 $\times$ 240) with a high noise level. 

\begin{table}[t]
\caption{\textbf{Demographics.} Distribution of height and weight for selected age and sex groups (mean $\pm$ 1 std dev). The age and sex composition are obtained by using the 2010 census data~\cite{AGESEX}, while the height and weight distributions are obtained by using the NHANES 1999-2002 census data~\cite{BODYWEIGHT}. This table was used to generate synthetic models using MakeHuman~\cite{MH}. Note that, unlike our approach, datasets such as CAESAR~\cite{CAESAR} do not truly  encompass the diversity of the population and are cumbersome and expensive to collect.}

\begin{tabular}{|l|l|l|l|l|l|l|l|l|l|}
\hline
\multicolumn{2}{|c|}{\multirow{4}{*}{Age}} & \multicolumn{8}{c|}{Gender} \\ \cline{3-10}
\multicolumn{2}{|c|}{} & \multicolumn{4}{c|}{Male} & \multicolumn{4}{c|}{Female} \\ \cline{3-10}
\multicolumn{2}{|c|}{} & \multicolumn{2}{c|}{\multirow{2}{*}{\begin{tabular}[c]{@{}c@{}}Height\\ (cm)\end{tabular}}} & \multicolumn{2}{c|}{\multirow{2}{*}{\begin{tabular}[c]{@{}c@{}}Weight\\ (kg)\end{tabular}}} & \multicolumn{2}{c|}{\multirow{2}{*}{\begin{tabular}[c]{@{}c@{}}Height\\ (cm)\end{tabular}}} & \multicolumn{2}{c|}{\multirow{2}{*}{\begin{tabular}[c]{@{}c@{}}Weight\\ (kg)\end{tabular}}} \\
\multicolumn{2}{|c|}{} & \multicolumn{2}{c|}{} & \multicolumn{2}{c|}{} & \multicolumn{2}{c|}{} & \multicolumn{2}{c|}{} \\ \hline
\multicolumn{2}{|l|}{18-24} & \multicolumn{2}{l|}{176.7$\pm$0.3} & \multicolumn{2}{l|}{83.4$\pm$0.7} & \multicolumn{2}{l|}{162.8$\pm$0.3} & \multicolumn{2}{l|}{71.1$\pm$0.9} \\ \hline
\multicolumn{2}{|l|}{25-44} & \multicolumn{2}{l|}{176.8$\pm$0.3} & \multicolumn{2}{l|}{87.6$\pm$0.8} & \multicolumn{2}{l|}{163.2$\pm$0.3} & \multicolumn{2}{l|}{75.3$\pm$1.0} \\ \hline
\multicolumn{2}{|l|}{45-64} & \multicolumn{2}{l|}{175.8$\pm$0.3} & \multicolumn{2}{l|}{88.8$\pm$0.9} & \multicolumn{2}{l|}{162.3$\pm$0.3} & \multicolumn{2}{l|}{76.9$\pm$1.1} \\ \hline
\multicolumn{2}{|l|}{65-74} & \multicolumn{2}{l|}{174.4$\pm$0.3} & \multicolumn{2}{l|}{87.1$\pm$0.6} & \multicolumn{2}{l|}{160.0$\pm$0.2} & \multicolumn{2}{l|}{74.9$\pm$0.6} \\ \hline
\end{tabular}
\label{table:Demographics}
\end{table}

In this paper, we propose a method to predict accurate body parameters and generate 3D human mesh models from a single depth map. We first create a large synthetic 3D human body model dataset using real-world body size distributions. Next, we extract body measurements from a single frontal-view depth map using joint location information provided by OpenNI \cite{OPENNI}. We combine estimates of body measurements with local geometry features around joint locations to form a robust multi-dimensional feature vector. This allows us to conduct a fast nearest-neighbor search to every 3D model in the synthetic dataset and return the closest one. Since the retrieved 3D model is fully parameterized and rigged, we can easily generate data such as standard full body measurements,labeled body parts, etc. Furthermore,we can animate the model by mapping the 3D model skeleton to joints provided by Kinect\texttrademark. Given the shape and pose parameters describing the body, we also developed an approach to fit a garment to the model with realistic wrinkles. A key benefit of our approach is that we only calculate simple features from the input data and search for the closest match in the highly-realistic synthetic dataset. This allows us to estimate a 3D body avatar in real-time, which is crucial for practical virtual reality applications. Also, compared to real-world human body datasets such as CAESAR \cite{CAESAR} and SCAPE \cite{SCAPE}, the flexibility of creating synthetic models enables us to represent more variations on body parameter distributions, while lowering cost and refraining from legal privacy issues involving human subjects. In summary, we make the following contributions:

\begin{itemize}
  \item We build a complete software system for body measurement extraction and 3D human model creation. The system only requires a single input depth map with real-time performance.
  \item We propose a method to generate a large synthetic human dataset following real-world body parameter distributions. This dataset contains detailed information about each sample, such as body parameters, labeled body parts and OpenNI-compatible skeletons. To our knowledge, we are the first to use large synthetic data that match distribution of true population for the purpose of model fitting and anthropometrics.
  \item We design a robust multi-dimensional feature vector and corresponding matching schemes for fast 3D model retrieval. Experiments show that our proposed method that uses simple computations on the input depth map returns satisfactory results. 

\end{itemize}

The remainder of this paper is organized as follows. Section~\ref{sec:Background} gives a summary of previous related work. 
Section~\ref{sec:ProposedSystem} discusses Im2Fit - our proposed system.  Section~\ref{sec:Experiments} shows our experimental results. Section~\ref{sec:Conclusion} provides some final conclusions and directions for future work.


\section{Background and Related Work}
\label{sec:Background}
We follow the pipeline to generate a large synthetic dataset, and then perform feature matching to obtain the corresponding 3D human model. A wealth of previous work has studied these two problems, and we mention some of them here to contrast with our approach.

{\bf{3D Human Body Datasets:}}
Despite vast literature concerning depth map based human body modeling, only a limited number of datasets are available for testing. These datasets typically contain real-world data to model the variation of human shape, and require a license to purchase.

The CAESAR~\cite{CAESAR} dataset contains few thousand laser scans of bodies of volunteers aged from 18 to 65 in the United States and Europe. The raw data of each sample consists of four scans from different viewpoints. The raw scans are stitched together into a single mesh with texture information. However, due to occlusion, noise and registration errors, the final mesh model is not complete.

The SCAPE~\cite{SCAPE} dataset has been widely used in human shape and pose estimation studies. Researchers found it useful to reshape the human body to fit 3D point cloud data or 2D human silhouettes in images and videos. It learns a pose deformation model from 71 registered meshes of a single subject in various poses. However, the human shape model is learned from different subjects with a standard pose. The main limitation of the SCAPE model is that the pose deformation model is shared by different individuals. The final deformed mesh is not accurate, since the pose deformation model is person-dependent. Also, the meshes are registered using only geometric information, which is unreliable due to shape ambiguities.

Recently, Bogo \etal~\cite{Bogo:CVPR:2014} introduced a new dataset called FAUST which contains 300 scans of 10 people with different poses. The meshes are captured by multiple stereo cameras and speckle projectors. All subjects are painted with high frequency textures so that the alignment quality can be verified by both geometry and appearance information.

\begin{figure}[!t]
\centering
\includegraphics[width=0.9\linewidth]{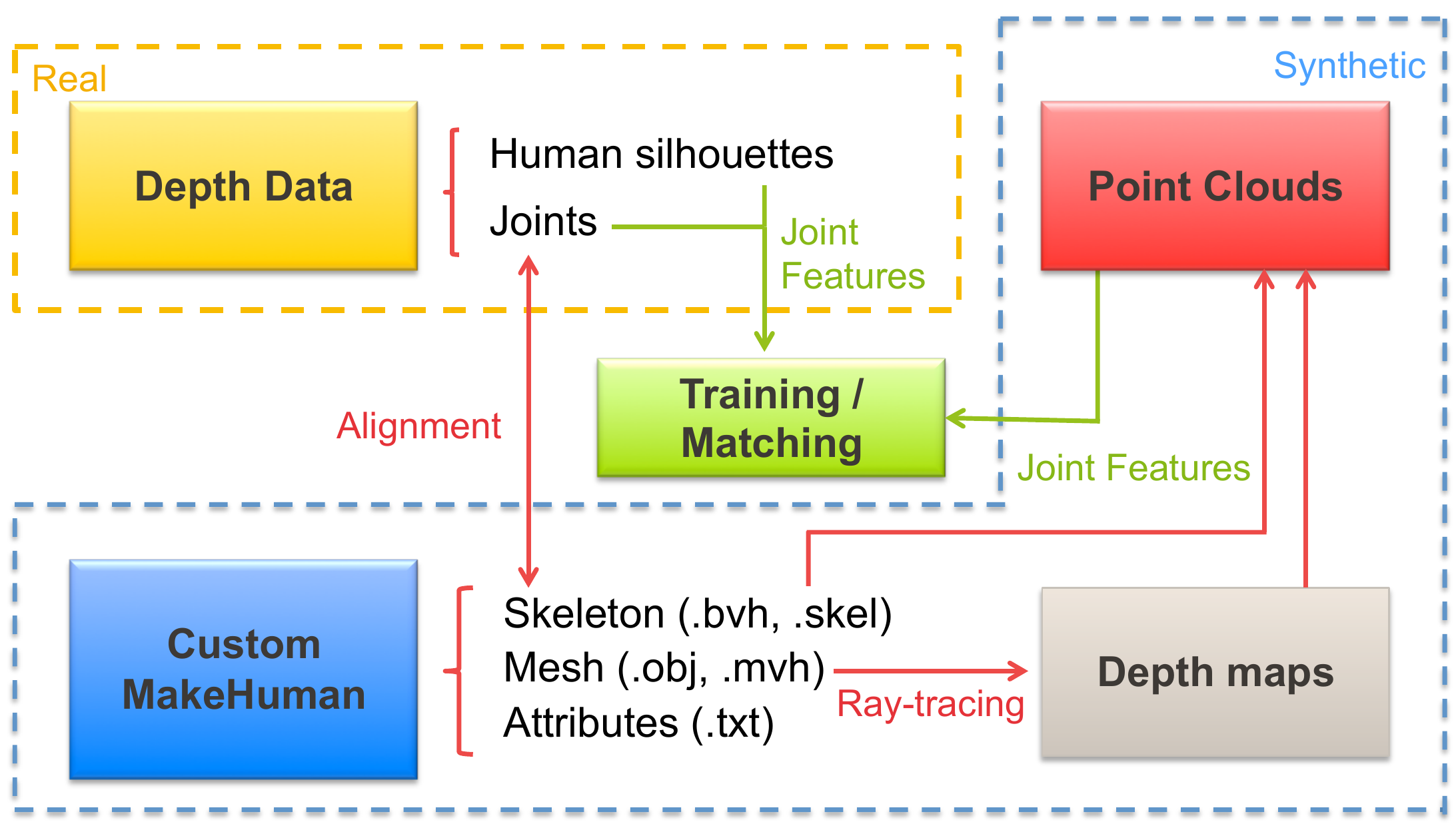}
\caption{\textbf{Block Diagram}. This illustrates the  interactions between various components of Im2Fit, along with relevant data. Details are in Section~\ref{sec:ProposedSystem}.}
\label{fig:BlockDiagram}
\vspace{-2ex}
\end{figure}

{\bf{Human Shape Estimation:}} A number of methods have been proposed to estimate human shape based on raw 3D data generated by a depth sensor or multiple images. Early work by Blanz and Vetter~\cite{Blanz:1999:MMS:311535.311556} use dense surface and color data to model face shapes. They introduced the term \emph{3D morphable model} to describe the idea of using a single generic model to fit all real-world face shapes. In their work, pose and shape are learned separately and then combined by using linear blend skinning (LBS). This inspires the SCAPE~\cite{SCAPE} model, which models deformation as  combination of a pose transformation learned from a single person with multiple poses and a shape transformation learned from multiple people with a standard pose. Balan \etal~\cite{balan2007detailed} fit the SCAPE model to multiple images. They later extend their work to estimate body shape under clothing~\cite{Balan:2008:NTE:1479250.1479253}. Guan \etal~\cite{guan2009estimating} estimated human shape and pose from a single image by optimizing the SCAPE model using a variety of cues such as silhouette overlap, edge distance and smooth shading. Hasler \etal~\cite{Hasler} developed a method to estimate 3D body shapes under clothing by fitting the SCAPE model to images using ICP~\cite{besl1992method} registration and Laplacian mesh deformation. Weiss \etal~\cite{Weiss11} proposed a method to model 3D human bodies from noisy range images from a commodity sensor by combining the silhouette overlap term, prediction error and an optional temporal pose similarity prior. More recently, researchers tackle the problem of decoupling shape and pose deformations by introducing a tensor based body model which jointly optimizes shape and pose deformations~\cite{Chen2013}.


\section{Im2Fit: The Proposed System }
\label{sec:ProposedSystem}
The overall processing pipeline of our software system can be found in Figure~\ref{fig:BlockDiagram}. Algorithm~\ref{DiscoveryAlgorithm} summarizes the various steps involved to estimate relevant anthropometrics for clothing fitment, once depth data and joint locations have been acquired.

\subsection{Depth Sensor Data Acquisition}
\label{sec:DataAcquisition}
\begin{figure*}[!ht]
\vspace{-1ex}
\begin{tabular}{ccc}
\includegraphics[width=0.15\linewidth]{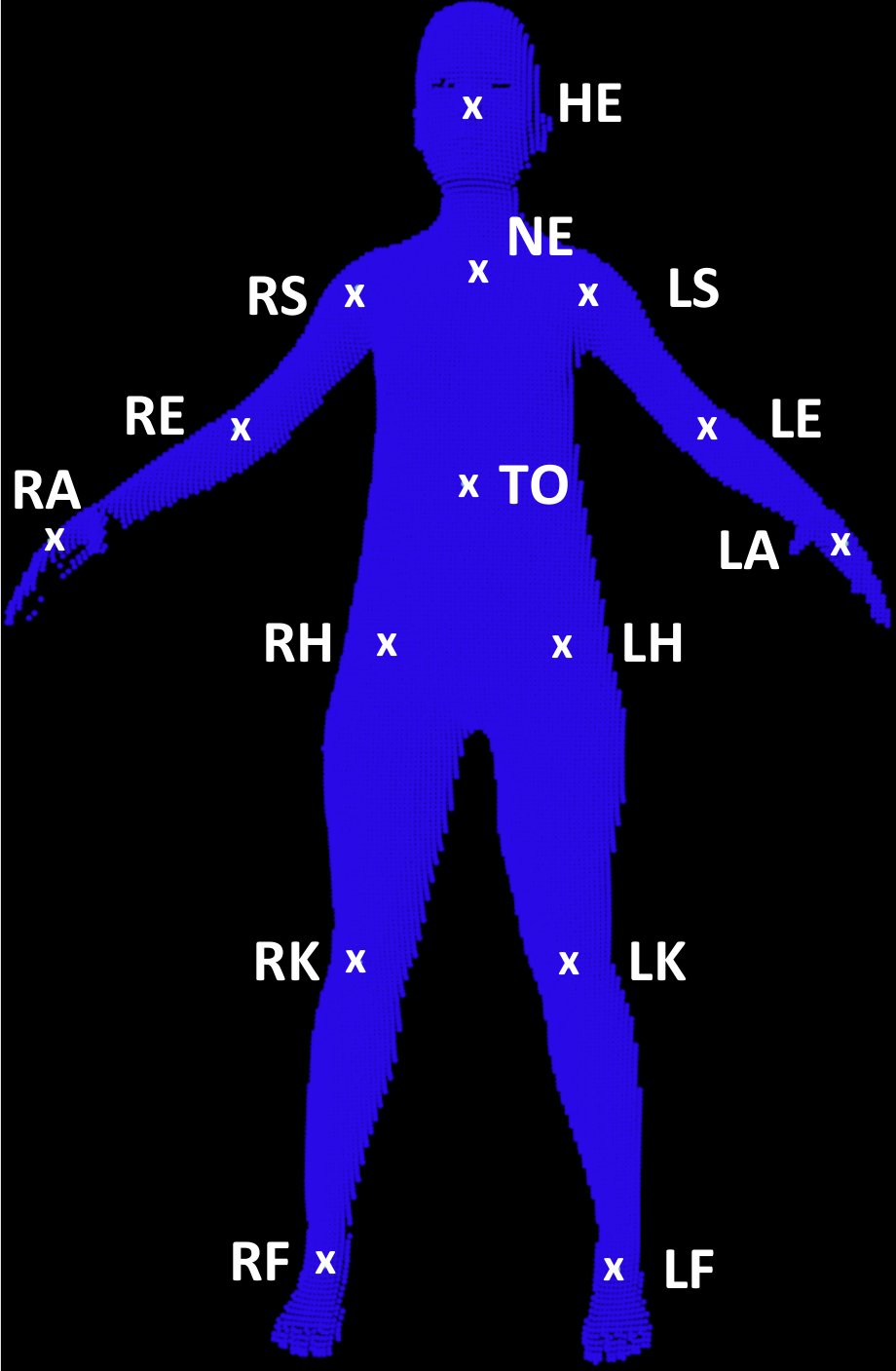} & 
\includegraphics[width=0.4\linewidth]{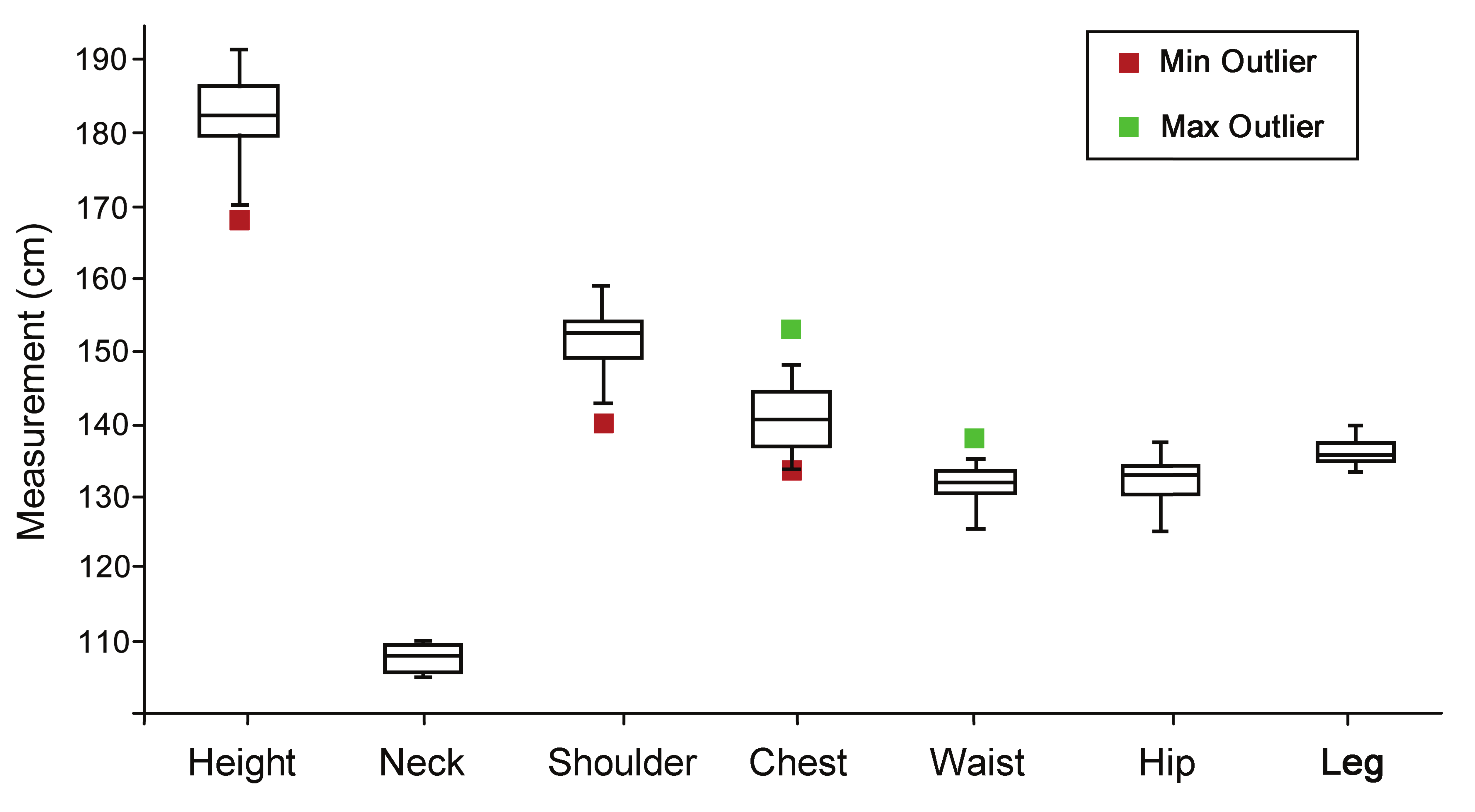}  &
\includegraphics[width=0.4\linewidth]{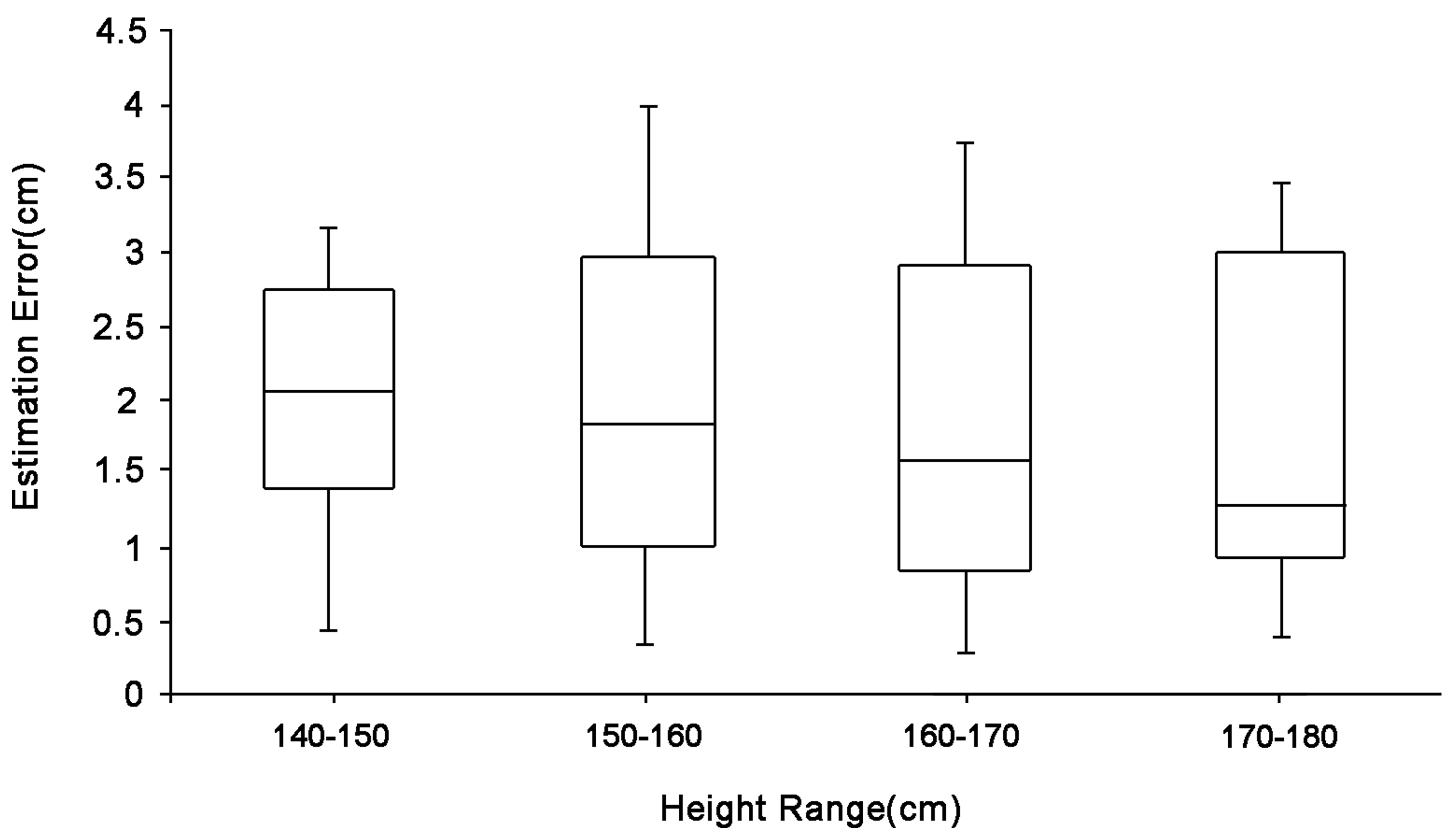} \\
\footnotesize{\hspace{0\linewidth}(a) Joints} & \footnotesize{\hspace{0\linewidth}(b) Measurement Distribution} &
\footnotesize{\hspace{0\linewidth}(c) Height Estimation Error} \\
\end{tabular}
\vspace{0em}
\caption{\textbf{Estimated Anthropometrics}. 
(a) We use 15 joints head, neck, torso and left/right hand, elbow, shoulder, hip, knee, foot, which we term \emph{HE, NE, TO, LA, RA, LE, RE, LS, RS, LH, RH, LK, RK, LF, RF}, respectively. (b) Distribution of key estimated measurements. We collected depth maps from 83 people and generated height, neck, shoulder, chest, waist, hip and leg measurements for each person. Plotted is the sample minimum, median and maximum, lower and upper quartile and outliers for different measurements.
(c) For a given height range, the distribution of height estimation errors are plotted. The overall mean estimation error for 83 samples is 1.87 cm.
 }
\label{fig:Anthropometrics}
\end{figure*}

The goal of real data acquisition process is to obtain user point cloud and joint locations, and extract useful features characterizing the 3D shape, such as body measurements and 3D surface feature descriptors. We first extract human silhouette from a Kinect\texttrademark~RGB-D frame and turn it into a 3D pointcloud. Next, we obtain joint locations by using OpenNI~\cite{OPENNI} which  provides a binary segmentation mask of the person, along with skeletal keypoints corresponding to different body joints. There are 15 joint positions in 3D real world coordinates (Figure~\ref{fig:Anthropometrics}(a)). They can be converted to 2D coordinates on the imaging plane by projection.

\begin{algorithm}
\caption{Frontal Measurement Generation}\label{DiscoveryAlgorithm}
\KwData{Depth Map $\bf{D}$ and estimated set of joints from pose estimation $\{j_{t}\}_{t=1,\ldots,T}$}
\KwResult{Set of Measurements $\textbf{M}$, where $t^{th}$ row is the measurement around joint $j_{t}$ }
\While{onEndCalibration=failure}{
   Call \emph{StartPoseDetection} \;
      \If{User detected}
      {
        Draw skeletons\;
        Compute principal axes $\left\{ \bf{u},\bf{v},\bf{w}\right\}$ \;
        Request calibration pose from user\;
        Call \emph{RequestCalibrationSkeleton}\;
         \If{User calibrated}
         { Save calibrated pose\;
           Call \emph{onEndCalibration}\;
                     \While{onEndCalibration=success}{
                 Call \emph{StartTrackingSkeleton} \;
                 \For{Joint label $t=1:T$}{
                  \If{$j_{t}$ is available}
                  {
                  Compute cross section plane.
                  \emph{/* defined by $\bf{v}$ and $\bf{w}$ passing through joint position $j_{t}$*/} \;
                  Compute intersection points\;
                  Ellipse fitting to obtain $m_{t}$\;
                  \If{$m_{t}$ is available for all $T$ joints}
                  {
                   Save $\textbf{M}$\;
                   Break\;
                  }
                  }
                  
                  }
                  }

         }
      }

}
\end{algorithm}

\subsection{Depth Sensor Data Processing - Measurements and Feature Extraction}
\label{sec:DataProcessing}
Once we obtain depth map and joint locations,, we generate features such as body measurements and Fast Point Feature Histograms (FPFH)\footnote{FPFH features are simple 3D feature descriptors to represent geometry around a specific point and are calculated for all joints.}~\cite{FPFH} around these joints. The body measurements include height, length of arm \& leg, girth of neck, chest, waist \& hip. These measurements can help us predict important user attributes such as gender, height, weight, clothing size. The aforementioned features are concatenated into a single feature vector for matching the features in the synthetic dataset.

We observe that the 6 points on the torso \emph{(NE, TO, LS, RS, LH, RH)} do not exhibit any relative motion and can be regarded on a whole rigid body. Therefore, we define the principal coordinate axes $\left\{ \bf{u},\bf{v},\bf{w}\right\}$ as $\bf{u}=\frac{\overrightarrow {\left( NE,TO\right) }} {\|\overrightarrow {\left( NE,TO\right) }\|},
\bf{v}=\frac{\overrightarrow {\left( LS,RS\right) }} {\|\overrightarrow {\left( LS,RS\right) }\|},
\bf{w}=\bf{u}\times\bf{v}$, as long as the following condition is satisfied:
\begin{equation}
\label{eq:PrincipalComponents}
\begin{split}
\frac{\bf{u} \cdot \overrightarrow {\left( TO,\frac {LH+RH} {2}\right) }} {\|\overrightarrow {\left( TO,\frac {LH+RH} {2}\right) }\|} < \varepsilon _{1}
\end{split}
\end{equation}
where $\overrightarrow {\left(A,B\right)}$ is the vector from point $A$ to point $B$, $\times$ is the vector cross-product, $\cdot$ is the vector dot-product, $\|\cdot\|$ is the Euclidean distance,  and $\varepsilon_1$ is a small positive fraction. We used $\epsilon_1=0.1$.   Condition~(\ref{eq:PrincipalComponents}) ensures that the posture is almost vertical. For simplicity, if this is not satisfied, we treat the data as unusable since error in measurement estimates will be large.

{\bf{Height:}} To calculate height, we first extract contour of the segmented 2D human silhouette, which was obtained by thresholding the depth map and projection on to 2D plane defined by $\bf{u}$ and $\bf{v}$. Next, we pick those points on the contour that satisfy the following condition:
\begin{equation}\label{eq:Height}
\frac {\left|  \bf{v} \cdot \overrightarrow {\left( TO,P_{c}\right)} \right|} {\left\| \overrightarrow {\left( TO,P_{c}\right)}\right\| } < \varepsilon_{2}
\end{equation}
where $P_{c}$ is an arbitrary point on the contour and $\varepsilon_2$ is a small positive fraction. We used $\varepsilon_2=0.1$. These 2D points lie approximately on $\bf{u}$. We sort them by their $y$ coordinates \footnote{Note that OpenNI uses GDI convention for the 2D coordinate system.} and find the top and bottom points. These points are converted to 3D real-world coordinates and the estimated height can be calculated as the Euclidean distance between the two points.

{\bf{Length of Sleeve and Legs:}} The sleeve length can be calculated as the average of $\left\|\overrightarrow{(LH,LE)}\right\|
+\left\|\overrightarrow{(LE,LS)}\right\|
+\left\|\overrightarrow{(LS,NE)}\right\|
$ and $\left\|\overrightarrow{(RH,RE)}\right\|
+\left\|\overrightarrow{(RE,RS)}\right\|
+\left\|\overrightarrow{(RS,NE)}\right\|
$. Note that these points are in 3D real-world coordinates. Similarly, length of legs can be obtained as the average of $\left\|\overrightarrow{(LH,LK)}\right\|
+\left\|\overrightarrow{(LK,LF)}\right\|
$
and
$\left\|\overrightarrow{(RH,RK)}\right\|
+\left\|\overrightarrow{(RK,RF)}\right\|
$.

{\bf{Girth of neck, shoulder, chest, waist and hip:}}
To get estimates of neck, shoulder, chest, waist and hip girth, we need to first define a 3D point $\bf{x}$, and then compute the intersection between the 3D point cloud and the plane passing through $\bf{x}$ and perpendicular to $\bf{u}$. Since the joints tracked by OpenNI are designed to be useful for interactive games, rather than being anatomically precise, we need to make some adjustments to the raw joint locations. Here we define new joint locations as follows:
\begin{equation}
\label{eq:NewPoints}
\begin{split}
\textbf{x}_{neck}=\frac {NE+HE} {2},
\textbf{x}_{shoulder}=\frac{\left( \textbf{x}_{neck}+\frac {LS+RS}{2}\right)}{2}\\
\textbf{x}_{chest}=\frac {NE+TO} {2},
\textbf{x}_{waist}=TO,
\textbf{x}_{hip}=\frac {LH+RH} {2}
\end{split}
\end{equation}
We can only get frontal view of the user, since the user is always directly facing the depth camera and we use a single depth channel as input. Therefore, we fit an ellipse~\cite{fitzgibbon1999direct} to the points on every cross-section plane to obtain full-body measurements. Such points are defined as 
$
\left\{ \textbf{p}\in\text{Point Cloud}, 
  s.t. \left | \frac{(\textbf{x}-\textbf{p})\cdot\bf{u}}{\|\textbf{x}-\textbf{p}\|} \right | < \varepsilon_3 \right\}
$, where $\varepsilon_3$ is a small positive fraction. We used $\varepsilon_3=0.1$.
Since our method is simple and computationally efficient, the measurements can be estimated in real time.

\subsection{Synthetic Data Generation  - Depth Map Rendering and Joint Remapping}
For realistic synthetic data generation, the goal is to create a large number of synthetic human data according to real world population distributions and save the feature vectors for each model. In order to generate the synthetic human model dataset, we developed a plug-in\footnote{We plan to make this plug-in available for research purpose and hope will flourish research in this area, which is stunted by the complexity of data acquisition for 3D human models and lack of royalty-free large public datasets.} for MakeHuman~\cite{MH}. 

MakeHuman is an open-source python framework designed to prototype realistic 3D human models. It contains a standard rigged human mesh, and can generate realistic human characters based on normalized attributes for a specific virtual character, namely Age, Gender, Height, Weight, Muscle and Ethnic origin (Caucasian, Asian, African). Our plug-in only modifies 4 attributes: Age, Gender, Height and Weight, which follows the distribution shown in Table~\ref{table:Demographics}. This is because we only care about the 3D geometry of the human body. Every synthetic model contains a Wavefront .obj file (3D mesh), a .skel skeleton file (Joint location), a Biovision Hierarchy .bvh (Rigged skeleton data) file and a text file containing the 9 above-mentioned attributes. We tested our plug-in on a desktop computer with Intel Xeon E5507 (4 MB Cache, 2.26 GHz) CPU and 12 GB RAM. The modified software is able to generate around 50,000 synthetic models in 12 hours (roughly a model every second). 

Since the synthetic data is clean, parametric and has fully labeled body parts and skeletons, we can match real world feature vectors extracted from depth frames directly to the MakeHuman synthetic feature vectors, and use the MakeHuman 3D parametric mesh model to replace incomplete and noisy point cloud data from consumer grade depth sensor. Since it is straightforward to establish joint correspondences, the 3D synthetic model can be deformed into any pose according to the pose changes of the user. The 3D mesh model can now be used for various applications such as virtual try-on, virtual 3D avatar or animation. Since it is difficult to directly process the MakeHuman mesh data, we also developed a scheme to render the mesh model into frontal and back depth maps and equivalent point clouds. The local 3D features can be extracted easily from these point clouds by using the Point Cloud Library (PCL)~\cite{PCL}.

The above process has several advantages. Firstly, our customized plug-in is able to generate a large number of realistic synthetic models with real world age, height, weight, or gender distribution, even dressed models with various clothing styles. Secondly, the feature vectors can be matched in real time. Thus, we can find the matching 3D model as soon as the user is tracked by OpenNI. Compared to methods such as \cite{Weiss11,Chen2013,barmpoutis2013} which try to generate parametric human model directly on the raw point cloud, our method is more robust and computationally efficient.

\subsection{Synthetic Data Retrieval  - Online 3D Model Retrieval}
Now that we obtained real-world and synthetic data in compatible formats, we can match feature vectors extracted from Kinect\texttrademark~depth frames directly to those generated from MakeHuman~\cite{MH}. The retrieved 3D human model can be used to replace the noisy, unorganized raw Kinect\texttrademark~point cloud. 

To reach the goal of real-time 3D model retrieval, we need to define a simple yet effective feature vector first. We divide the feature vector into three groups with different weights:  global body shape, gender and local body shape. 

The first group contains 4 parameters: height and length of sleeve, leg and shoulder. This captures the longer dimensions of the human body. 

The second group contains two ratios. They are defined as follows:
\begin{equation}
ratio_1=\dfrac { \bigfrown{\left\| Chest\right\|} } {\overrightarrow {\left\| Chest\right\| }}
, ratio_2=\dfrac {Hip}{Waist}
\end{equation}
where $\bigfrown{\left\| Chest\right\|}$  represents the 3D surface geodesic distance and $ \overrightarrow {\left\| Chest\right\| }$ represents the Euclidean distance between point LS and RS. If two body shapes are similar, these two ratios tend to be larger for females and smaller for males. 

The third group contains FPFH (33 dimensions) features~\cite{FPFH} computed at all 15 joints with a search radius of 20 centimeters. This describes the local body shape. The feature vector has $4+2+33\times15=501$ dimensions. Any pair of feature vectors can be compared simply by using their $L_2$ distance.

We use nearest neighbor search to find the closet match in the synthetic dataset. The size of the feature vector dataset for 50,000 synthetic models is roughly 25 MB, and a single query takes about 800 ms to complete on a 2.26 GHz Intel Xeon E5507 machine.



\section{Experiments}
\label{sec:Experiments}
We conducted our experiments on clothed subjects with various body shapes. Figure \ref{fig:Anthropometrics}(b) shows the distribution of measurements from 83 people predicted by our system. For privacy issues, we were only able to get ground truth height, gender and clothing size data. 

\begin{figure*}[ht]
\centering
\includegraphics[width=0.9\linewidth]{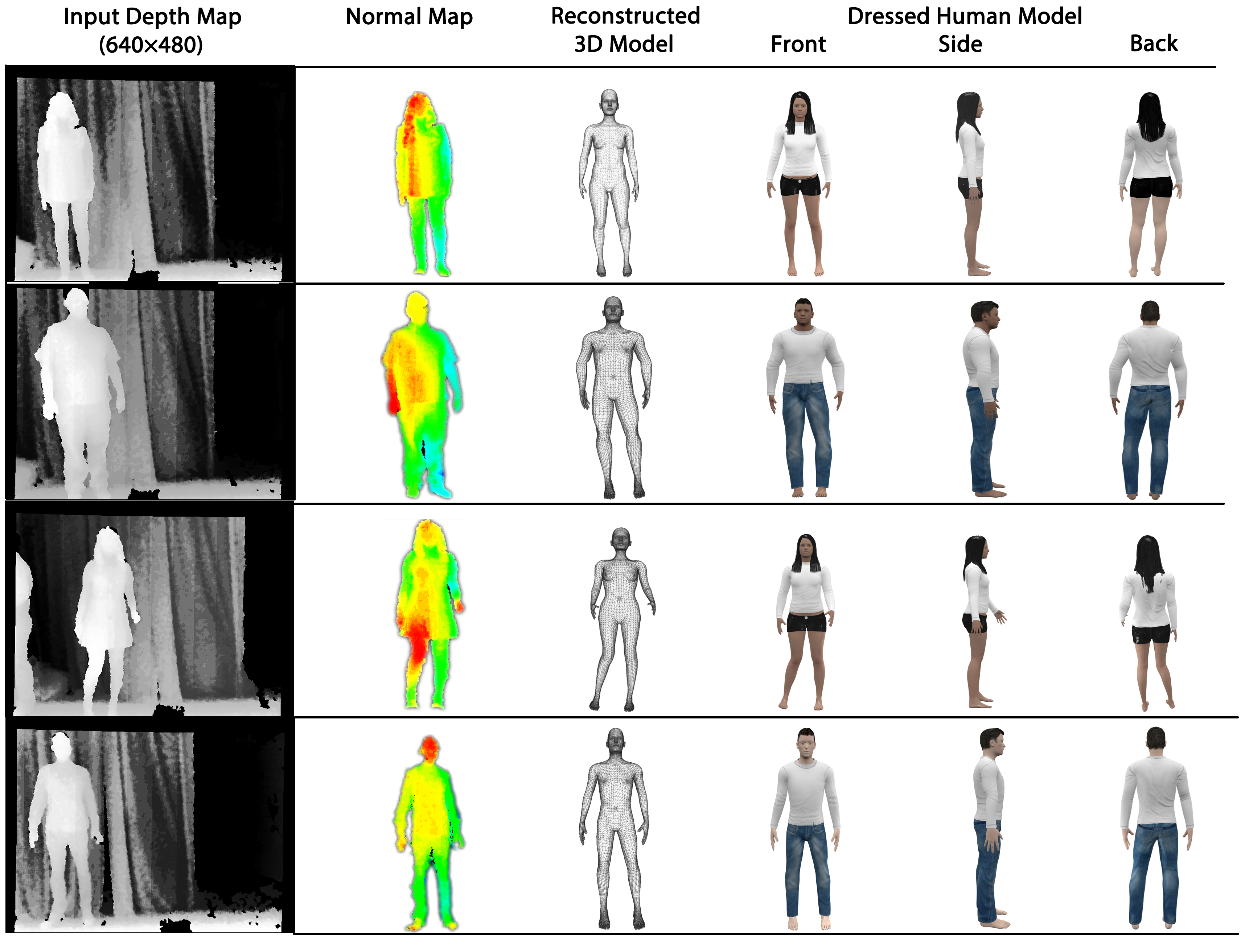}
\caption{\textbf{Qualitative Evaluation}. The first column shows the raw depth map from Kinect\texttrademark~360 sensor, the second column shows the normal map which we use to compute the FPFH~\cite{FPFH} features. The third column shows the reconstructed 3D mesh model, and the last three columns show the front, side and back view of the 3D model after basic garment fitting.}
\label{fig:Qualitative}
\vspace{-2ex}
\end{figure*}

\begin{figure}[h]
\centering
\includegraphics[width=0.9\linewidth]{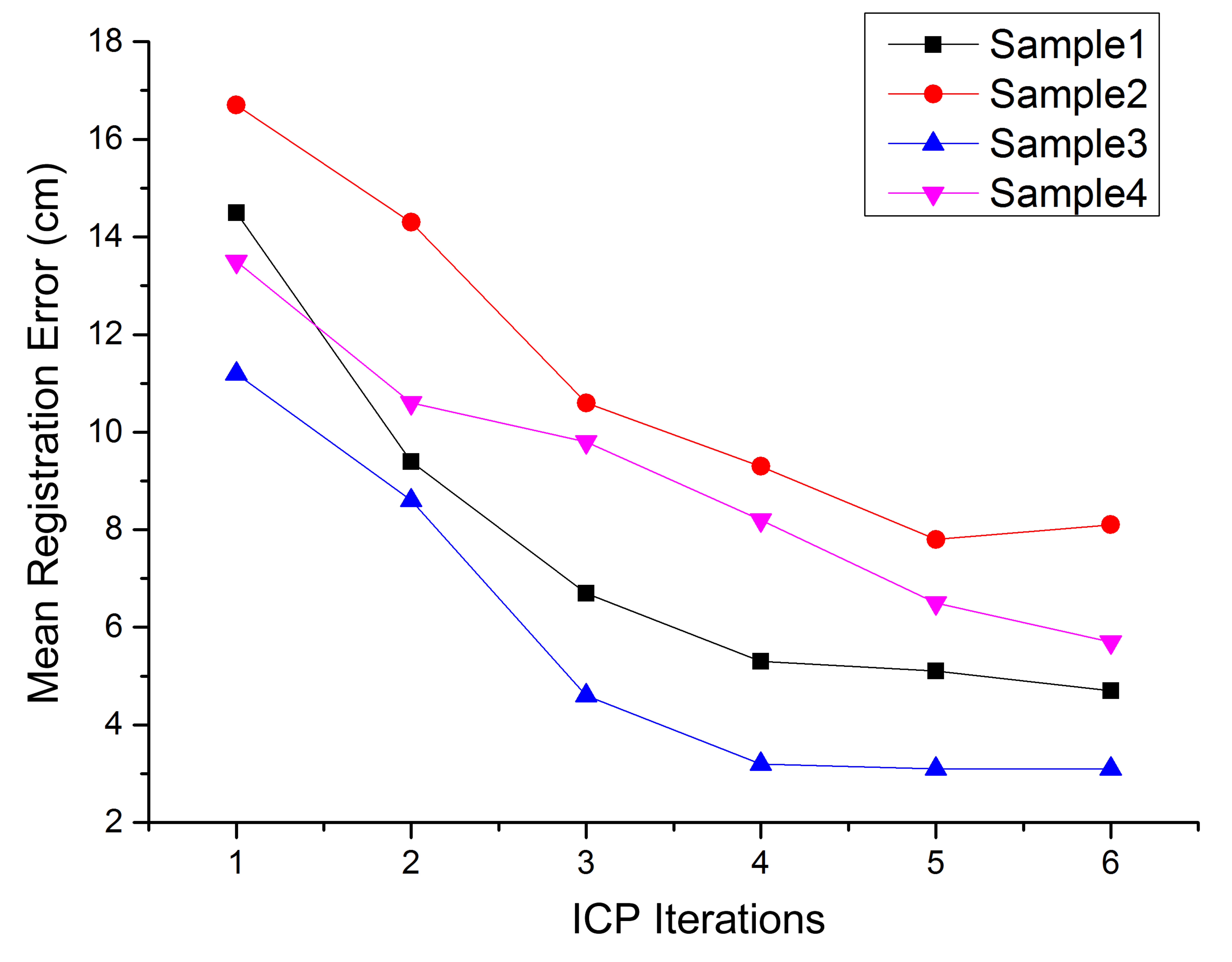}
\caption{\textbf{ICP Error}. Mean registration error per ICP~\cite{besl1992method} iteration for different samples. We evaluated ICP registration error for 4 samples in the dataset. The raw depth maps and estimated body shapes for the 4 samples can be found in Figure \ref{fig:Qualitative}. 
}
\label{fig:ICPError}
\vspace{-2ex}
\end{figure}

The gender predicted by $Ratio_{1}$ and $Ratio_{2}$ is correct for all 83 samples. The height estimation error for different height ranges is shown in Figure \ref{fig:Anthropometrics}(c). The overall mean estimation error for the 83 samples is 1.87 centimeters. Compared to \cite{Weiss11}, our system has obtained competitive accuracy with a much simpler algorithm. The results of our system can be qualitatively evaluated from Figure \ref{fig:Qualitative}. To verify the prediction accuracy, we applied the Iterative Closest Point (ICP)~\cite{besl1992method} registration algorithm to obtain the mean geometric error between the retrieved 3D mesh model and the raw point cloud data for the four subjects shown in Figure \ref{fig:Qualitative}. The results are shown in Figure \ref{fig:ICPError}. We simply aligned the skeleton joints for ICP initialization. After 6 ICP iterations, the mean registration error converges to below 9 centimeters, for all four subjects. This is because the feature vector gives more weights to frontal measurements.  The mesh alignment can be fine tuned by ICP after several iterations. However, the initial joint locations are only accurate on the 2D plane, and may shift in the $z$-direction due to random noise or holes on the point cloud surface. 
\begin{figure}[!t]
\centering
\includegraphics[width=0.9\linewidth]{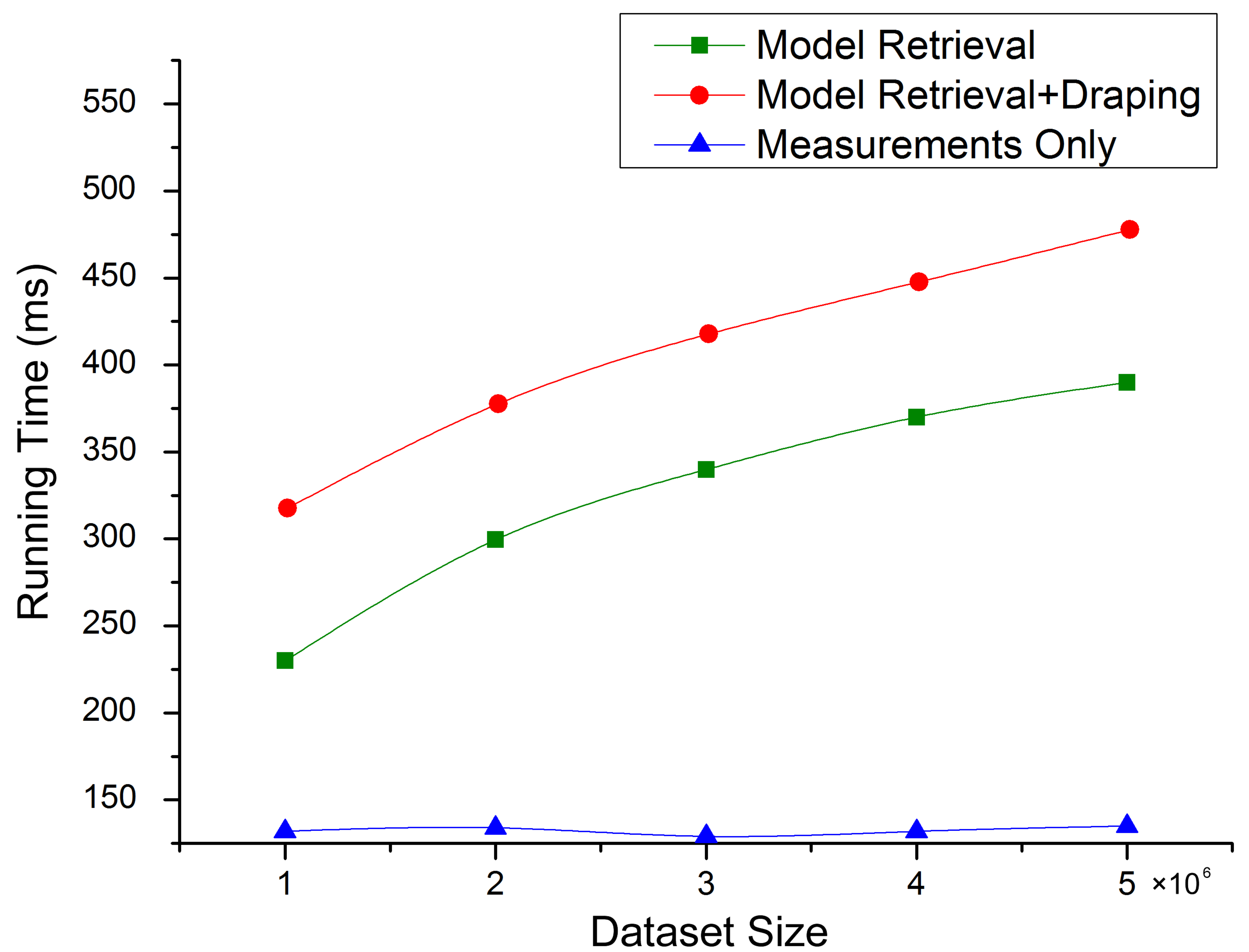}
\caption{\textbf{Runtime}. Runtime of our complete system with increasing size of the synthetic dataset. For synthetic dataset with given number of samples, we evaluate the average runtime end-end, to estimate relevant measurements, naked 3D models and clothed 3D models.
}
\label{fig:Runtime}
\vspace{-2ex}
\end{figure}


Runtime of 3D model retrieval algorithm with different size of the synthetic dataset is summarized in Figure~\ref{fig:Runtime}. The algorithm was run on a single desktop without GPU acceleration to obtain relevant measurements, naked body model and clothed model. The measurement estimation only depends on the input depth map. So the running time is almost constant on a depth map with given resolution. Generating clothed models is slower than naked models because of the garment fitting and rendering overhead which can be improved by GPU acceleration. On a dataset containing $5\times10^{6}$ samples, we achieved runtime of less than 0.5 seconds for a single query, with little memory usage. Compared to \cite{Weiss11} which takes approximately 65 minutes to optimize, our method is significantly faster, while still maintaining competitive accuracy. 

Using measurements from the body meshes, we also predicted clothing sizes and compared with ground truth data provided by the participants. The overall clothing size prediction accuracy is $87.5\%$. We estimated size of T-shirt (\eg XS, S, M, L, XL, 2XL, 3XL).

\section{Conclusion}
\label{sec:Conclusion}
We presented a system for human body measurement and modeling using a single consumer depth sensor, using only the depth channel. Our method has several advantages over existing methods. First, we propose a method for generating synthetic human body models following real-world body parameter distributions. This allows us to avoid complex data acquisition rig, legal and privacy issues involving human subject research, and at the same time create a synthetic dataset which represents more variations on body shape distributions. Also, models inside the synthetic dataset are complete and clean (\ie no incomplete surface), which is perfect for subsequent applications such as garment fitting or animation. Second, we presented a scheme for real-time 3D model retrieval. Body measurements are extracted and combined with local geometry features around key joint locations to form a robust multi-dimensional feature vector. 3D human models are retrieved by using a fast nearest neighbor search, and the whole process can be done within half a second on a dataset containing $5\times10^{6}$ samples. Experiment results have shown that our system is able to generate accurate results in real-time, thus is particularly useful for home-oriented body scanning applications on low computing power devices such as depth-camera enabled smartphones or tablets.

\section{Appendix - Implementation Details}
\label{sec:ImplementationDetails}

\subsection{Depth Sensor Data Acquisition - Skeleton Detection}
\label{sec:DataAcquisition}
The Kinect\texttrademark~depth sensor provides two sources of data, namely the RGB and depth channels. This work purely focusses on the depth channel, which can be used to infer useful information about the person's 3D location and motion. We utilize open source SDKs to detect and track the person of interest. The SDK provides a binary segmentation mask of the person, along with skeletal keypoints corresponding to different body joints.

\subsection{Depth Camera SDKs} 

Compared to the Kinect\texttrademark~SDK which only works on Windows Platforms, OpenNI is more popular in open source and cross-platform projects. Although both SDKs provide access to basic RGB images and depth maps, the human silhouette tracking and joint location estimation are implemented in different algorithms. According to \cite{Cosgun2013}, OpenNI performs better for head and neck tracking, whereas Microsoft\textregistered~Kinect\texttrademark~SDK is more accurate on other joints. From our experience, the Microsoft\textregistered~Kinect\texttrademark~SDK outperforms OpenNI in both joint location estimation and human silhouette tracking and segmentation.

The Kinect\texttrademark~360 sensor utilizes infrared structured light to perceive depth. It has a resolution of $640\times480$ at 30 frames per second, with working range of 0.8m to 4m. In practice, a minimum distance of 2m is required for tracking the full body. The field of view is 57 degrees horizontal, 43 degrees vertical. The Kinect\texttrademark~device also provides a tilting motor which operates from -27 to 27 degrees with an accelerometer for detection the tilting angle. This is originally designed for automatically tracking the ground plane. The hardware specifications of the Kinect\texttrademark~sensor can be found in table 1.

\begin{figure}[!t]
\centering
\includegraphics[width=0.9\linewidth]{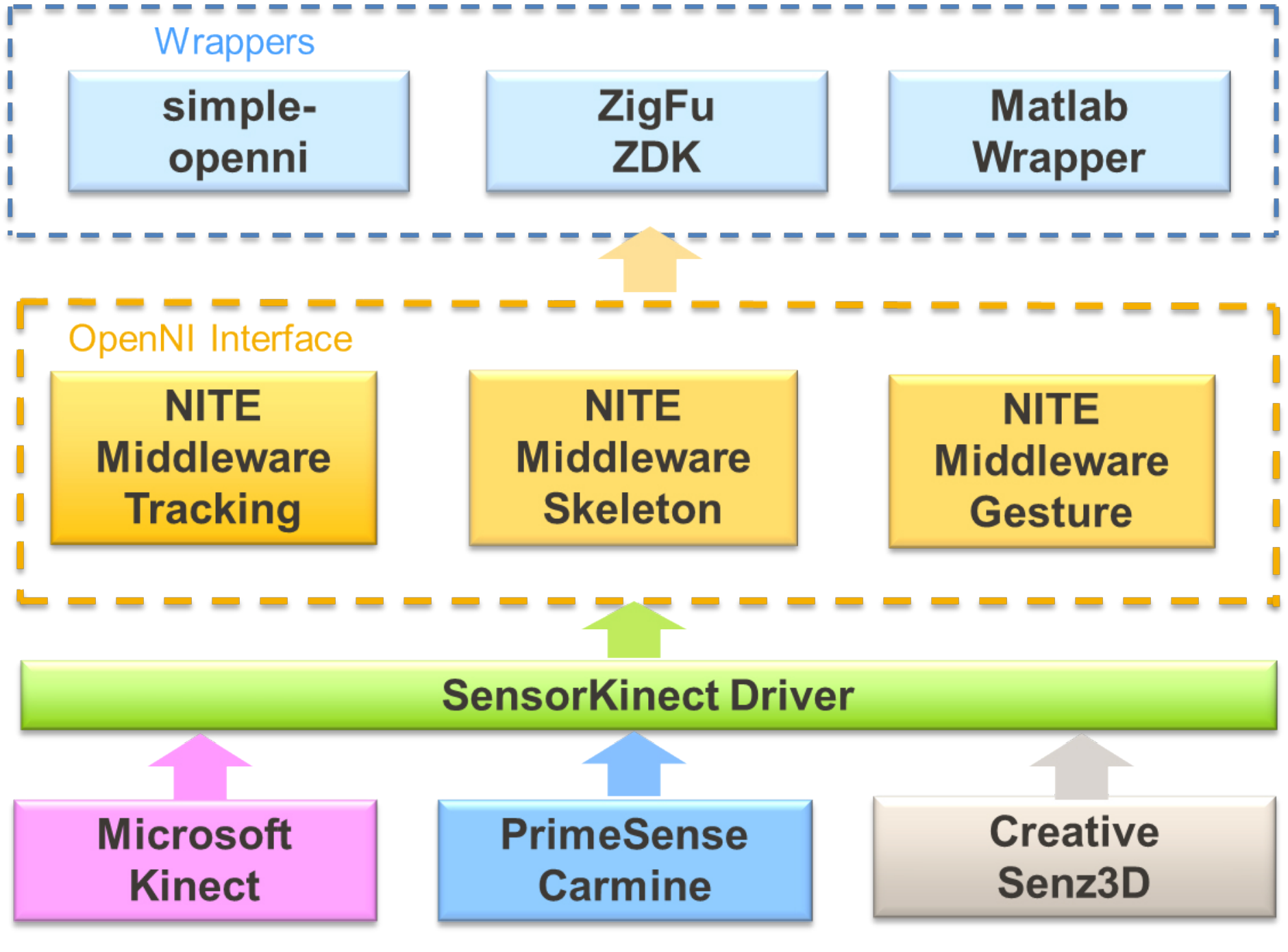}
\caption{\textbf{Platform Stack.} See Section~\ref{sec:DataAcquisition} for details.}
\label{fig:Platform}
\vspace{-2ex}
\end{figure}

\begin{figure*}[!ht]
\centering
\includegraphics[width=0.9\linewidth]{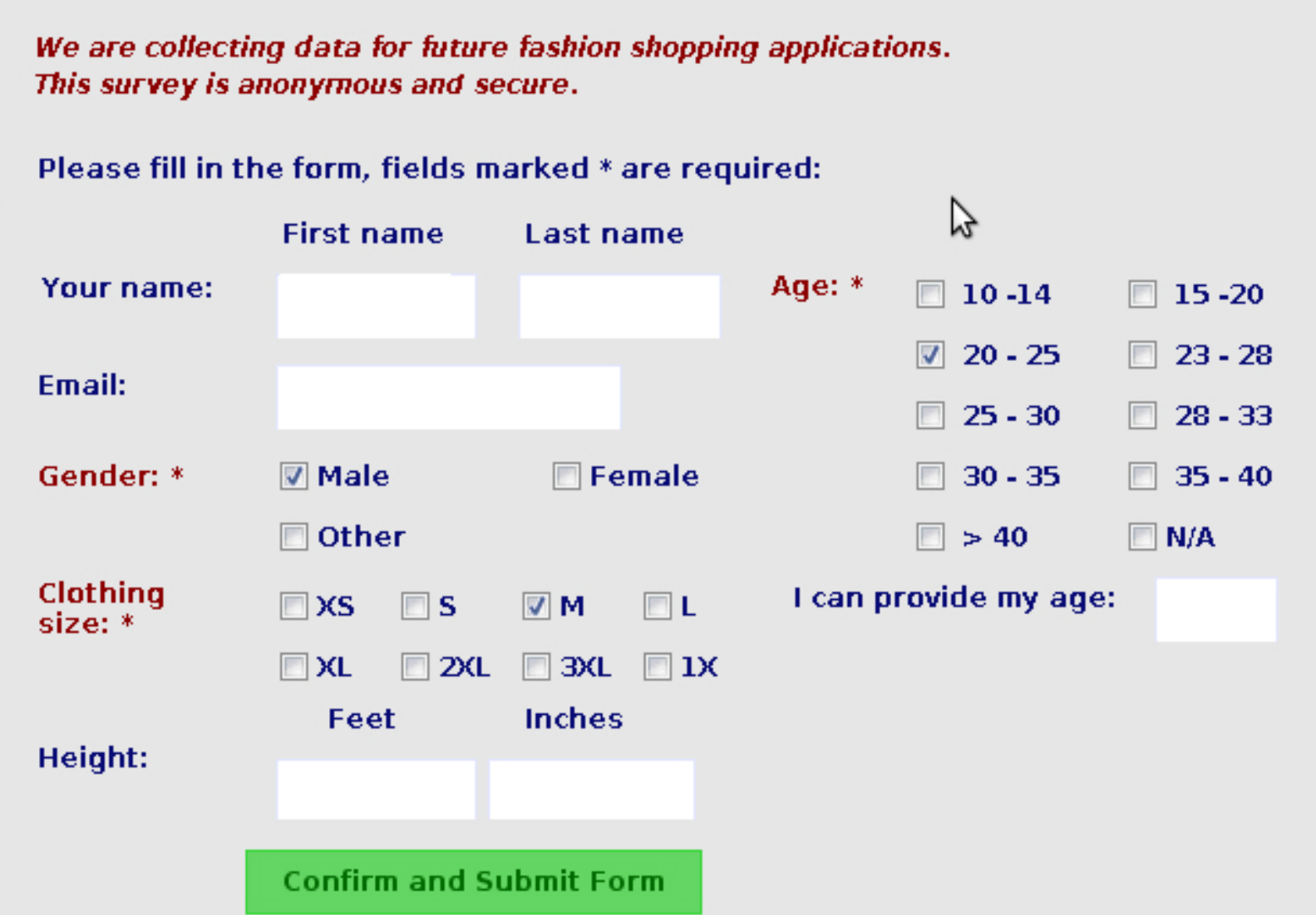}
\caption{\textbf{User Interface.} This shows part of the user interface used to collect data from subjects using consumer grade depth sensor.}
\label{fig:UserInterface}
\vspace{-2ex}
\end{figure*}

\subsection{Data Acquisition}
\label{sec:DataAcquisition}
We choose the OpenNI framework as the SDK for data acquisition. Our graphical user interface (GUI) (see Figure~\ref{fig:UserInterface}) is able to track the user and record joint locations in real time. Also, a form is provided on a separate window for the user to enter attributes such as height, weight, age, gender and so on as ground truth data.  Besides the OpenNI SDK, our software requires two additional packages, namely SensorKinect\texttrademark~driver and NITE middleware. The SensorKinect\texttrademark~provides low level drivers for accessing raw color image, depth map and infrared camera images from the Kinect\texttrademark~sensor. The NITE middleware, on the other hand, provides various features such as human silhouette segmentation, tracking, gait analysis and skeletonization. The OpenNI SDK is implemented in C++. Additionally, there are popular wrappers for the OpenNI such as simple-openni for Java \cite{SIMPLEOPENNI}, ZigFu ZDK for Unity 3D engine, Adobe Flash and HTML5 \cite{ZIGFU}, and an official wrapper for Matlab \cite{MATKIN}. These wrappers make it easier for programmers to call certain OpenNI functions using various programming languages. Figure ~\ref{fig:Platform} shows the relationship between the aforementioned components.

\subsection{Virtual Human Model Generation}
\label{sec:ModelGeneration}
We showed that it is possible to automatically create a large-scale synthetic human database following real-world distributions. However, the synthetic data is still not compatible with data obtained from the Kinect\texttrademark~device, for the following reasons:
\begin{itemize}
  \item The Wavefront .obj is in mesh format, which represents the 3D geometry in vertices and faces. For Kinect, the data is in color images and depth maps (or RGBXYZ point clouds).
  \item The joints generated by MakeHuman~\cite{MH} are from the built-in game.json 32 bone skeleton system, which is not compatible with the OpenNI skeleton system with 15 joints.
\end{itemize}
To address these problems, we developed an additional program to process the data generated by MakeHuman, using OpenGL and OpenCV. This program renders a frontal and back depth map of the MakeHuman model with similar depth sampling rate to the Kinect\texttrademark~sensor and remaps the game.json joints to 15 OpenNI-compatible joints. The detailed process is discussed in Section~\ref{sec:DataAcquisition}.

The OpenNI SDK provides default camera models with fairly accurate focal lengths for the \emph{convertRealWorldToProjective} and \emph{convertProjectiveToRealWorld} function. However, the camera model is only assumed to be a simple pinhole model without lens distortion. This is because Kinect\texttrademark~uses low-distortion lenses, so the maximum distortion error will be only a few pixels. However, our application requires maximum accuracy from the Kinect's 3D data, so we still performed a standard camera calibration and saved the intrinsic parameters as an initialization .yaml file, which is used for both depth map rendering and joint remapping.

To render the depth map, we assume that the virtual camera is placed 2 meters in front of the 3D human model. We provide the customized projection matrix to the OpenGL Z-buffering routine to render the 8-bit unsigned integer depth map. The depth maps are stored in both high resolution and low resolution formats. The size of the low resolution depth maps is set to \begin{math}640\times480\end{math} to simulate the Kinect\texttrademark~sensor.

The joint remapping from MakeHuman to OpenNI is simple, as MakeHuman skeleton system contains all joints that OpenNI has. The 3D coordinates of MakeHuman game.json file are then projected to the imaging plane using the aforementioned projection matrix to align with the rendered depth map.

{\small
\bibliographystyle{unsrt}
\bibliography{Im2Fit}
}

\end{document}